\documentclass[runningheads]{llncs}

\usepackage[mobile]{eccv}

\usepackage{eccvabbrv}

\usepackage{graphicx}
\usepackage{booktabs}
\usepackage{algorithm}
\usepackage{algpseudocode}
\usepackage{amsmath}
\usepackage{amssymb}
\usepackage{needspace}
\usepackage{wrapfig}
\usepackage{gensymb}

\usepackage{amsmath,amsfonts,bm}

\def\eqref#1{equation~\ref{#1}}

\def\1{\bm{1}}

\DeclareMathAlphabet{\mathsfit}{\encodingdefault}{\sfdefault}{m}{sl}
\SetMathAlphabet{\mathsfit}{bold}{\encodingdefault}{\sfdefault}{bx}{n}

\usepackage{url}
\usepackage{lineno}
\usepackage{booktabs}
\usepackage{lipsum}
\usepackage{graphicx}
\usepackage{float}
\usepackage{amsmath,amsfonts,amssymb}
\usepackage{xcolor}
\usepackage{colortbl}
\usepackage{caption}
\usepackage{wrapfig}
\usepackage{caption} 
\usepackage{comment}

\renewcommand{\paragraph}[1]{\vspace{.5em}\noindent\textbf{#1}}

\usepackage[shortlabels,inline]{enumitem}
\setlist[itemize]{noitemsep,leftmargin=*,topsep=0em}
\setlist[enumerate]{noitemsep,leftmargin=*,topsep=0em}

\newcommand{\image}{I}
\newcommand{\timestep}{t}
\newcommand{\lastime}{T}
\newcommand{\view}{p}
\newcommand{\numberg}{N}
\newcommand{\Gaussian}{\mathbf{G}}
\newcommand{\centr}{\mu}
\newcommand{\quat}{q}
\newcommand{\scale}{s}
\newcommand{\opacity}{o}
\newcommand{\colour}{c}
\newcommand{\rank}{r}
\newcommand{\func}{F_\phi}

\newcommand{\feat}{\mathbf{z}}

\newcommand{\mlp}{\psi}

\definecolor{tabfirst}{rgb}{1, 0.7, 0.7} 
\definecolor{tabsecond}{rgb}{1, 0.85, 0.7} 
\definecolor{tabthird}{rgb}{1, 1, 0.7} 
\definecolor{darkyellow}{rgb}{0.85,0.65,0.0}

\usepackage[accsupp]{axessibility}  

\usepackage{hyperref}

\usepackage{orcidlink}

\begin{document}

\newcommand{\methodname}{\textsc{GrowFlow}\xspace}
\title{Grow with the Flow: 4D Reconstruction of Growing Plants with Gaussian Flow Fields}
\titlerunning{GrowFlow}

\author{Weihan Luo$^{1 }$\thanks{weihan262144@outlook.com} \quad
Lily Goli$^{1,2}$ \quad
Sherwin Bahmani$^{1,2}$ \\
Felix Taubner$^{1,2}$ \quad
Andrea Tagliasacchi$^{1, 3}$ \quad
David B. Lindell$^{1, 2}$ \vspace{-0.5em}}

\authorrunning{W. ~Luo et al.}

\institute{$^1$University of Toronto \quad $^2$Vector Institute \quad $^3$Simon Fraser University \\ {\small\href{https://weihanluo.ca/growflow/}{weihanluo.ca/growflow/}} \vspace{-1.5em}} 

\maketitle

\begin{abstract}
Modeling the time-varying 3D appearance of plants during growth poses unique challenges: unlike most dynamic scenes, plants continuously generate new geometry as they expand, branch, and differentiate.
Existing dynamic scene representations are ill-suited to this setting: deformation fields provide insufficient constraints to yield physically plausible scene dynamics, and 4D Gaussian splatting represents the same physical structures with different Gaussian primitives at different times, breaking temporal consistency.
We introduce \methodname, a dynamic representation that couples 3D Gaussian primitives with a neural ordinary differential equation to model plant growth as a continuous flow field over geometric parameters (position, scale, and orientation).
Our representation enables consistent appearance rendering and models nonlinear, continuous-time growth dynamics with full temporal correspondences for every primitive.
To initialize a sufficient set of Gaussian primitives, we first reconstruct the mature plant and then learn a reverse-growth process, effectively simulating the plant's developmental history in reverse.
\methodname achieves superior image quality and geometric coherence compared to prior methods on a new, multi-view timelapse dataset of plant growth, and provides the first temporally coherent representation for appearance modeling of growing 3D structures.
\end{abstract}

\section{Introduction}
\label{sec:intro}
Accurately modeling plant growth has wide-reaching implications for plant phenotyping, agriculture, and biological research, where understanding the temporal development of plant structures is essential for analyzing morphology, function, and environmental response~\cite{dhondt2013cell, pound2017deep, rincon2022four, owens2016modeling, ijiri2014flower}. Unlike most dynamic scenes, plant growth is inherently non-rigid and involves continuous structural change: new leaves and branches emerge gradually, altering both geometry and topology over time~\cite{coen2023mechanics, sinnott1960plant, li2013analyzing, geng2025birth, wang2025autoregressive}. 
We address the problem of reconstructing time-varying 3D representations of plant growth from multi-view time-lapse imagery, with a particular focus on capturing temporally coherent geometry throughout development.
\begin{figure}[t!]
\centering
\includegraphics[width=\textwidth]{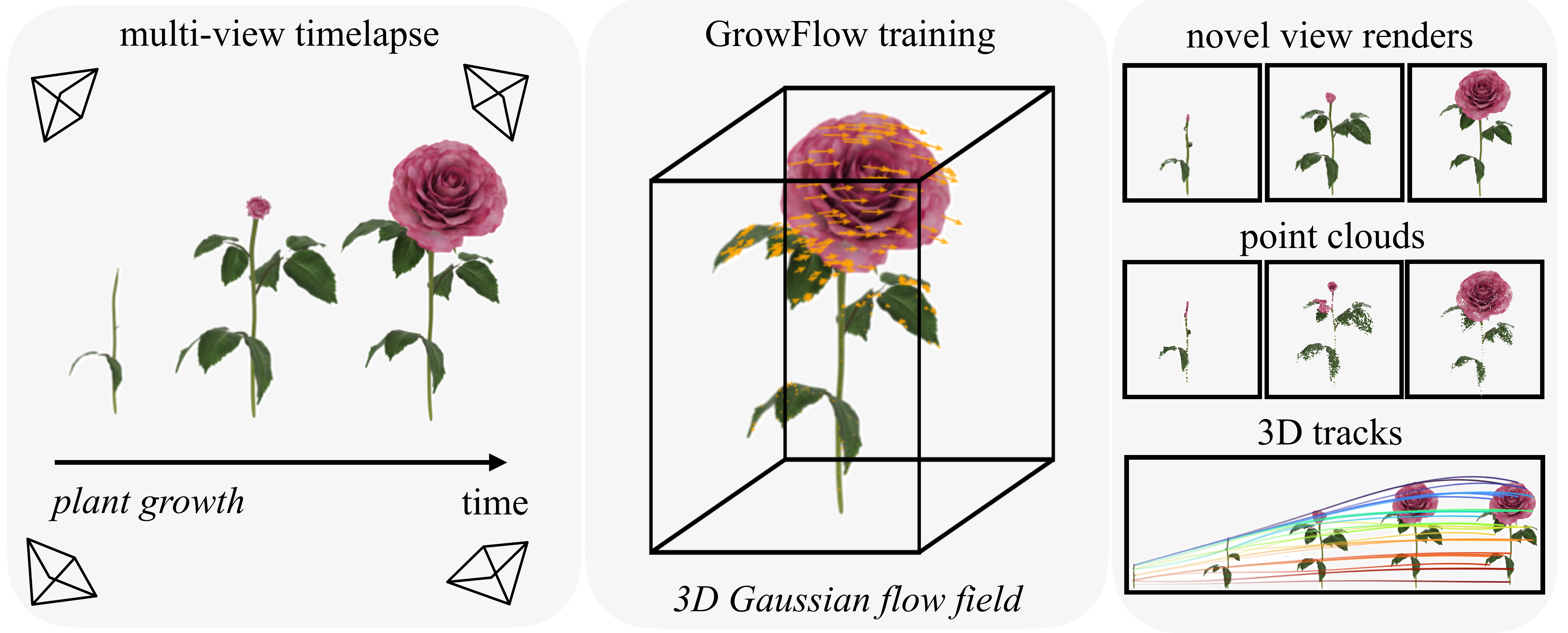}
\captionof{figure}{\textbf{\methodname.} We propose \methodname, a method for 
reconstructing high-fidelity geometry of plant growth. Given multi-view timelapse 
images of a plant, our method accurately reconstructs the dynamic structure using 
a set of 3D Gaussian primitives and a flow field defined over their parameters. 
Our continuous flow field further enables temporal interpolation of both geometry 
and appearance between frames. We can also track structures during a plant's growth 
by visualizing the positions of the 3D Gaussian primitives, as shown above for the synthetic \textit{rose} plant. Please see the Supp.\ Webpage for video results.
}
\label{fig:teaser}
\end{figure}

Contemporary dynamic scene representations fall broadly into two families, neither of which is well-suited to our problem.
Deformation-based methods~\cite{wu20244d, yang2024deformable} map a canonical representation to scene structure at each timestep via a learned deformation field, but impose little constraint on the smoothness or physical plausibility of the field---nothing prevents learning geometrically implausible mappings that merely minimize the photometric loss.
Methods based on 4D Gaussians with temporal masking~\cite{li2024spacetime, duan20244d, yang20244d} are even less constrained: geometry is discarded and introduced across time with no notion of correspondence.
Most closely related to our setting, 
GrowSplat~\cite{adebola2025growsplat} applies 3D Gaussian Splatting (3DGS)~\cite{kerbl20233d} to plant growth, but produces independent per-timestep reconstructions that similarly lack temporal correspondences.
In growth modeling, tracking the development of individual leaves and branches over time is as important as rendering quality---and previous work does not meet this requirement.

We propose a new perspective: plant growth can be modeled as a continuous dynamical system, where each scene element follows a smooth trajectory through space and time, governed by an underlying vector field.
We parameterize this vector field as a neural ordinary differential equation (ODE)~\cite{chen14torchdiffeq}, whose integration naturally enforces smooth, continuous evolution, as the Gaussian trajectories are constrained to follow a consistent vector field---providing an inductive bias that unconstrained deformation fields lack.

Building on this insight, we present \methodname, a novel dynamic representation that couples 3D Gaussian primitives with a neural ODE to learn this growth vector field, yielding a temporally coherent and biologically plausible evolution of plant geometry, as shown in Fig.~\ref{fig:teaser}.
A key challenge in this setting is how to continuously introduce new geometry as the plant grows: directly adding new Gaussians is non-differentiable and hard to optimize.
We sidestep this by reconstructing the mature plant and learning growth in reverse—modeling the plant's developmental history backwards through time.
Concretely, we learn a continuous ODE flow field over the position, scale, and orientation of 3D Gaussian primitives, while keeping color and opacity fixed, then reverse this process to recover a realistic growth trajectory.
Because all Gaussians persist throughout the ODE trajectory, each primitive maintains a consistent identity across time, enabling the kind of geometric coherence that existing methods cannot provide.
While this restricts \methodname to monotonic growth, where plant structure only accumulates over time, this assumption holds broadly in plant phenotyping and agricultural settings, where \methodname achieves state-of-the-art performance in both novel-view and novel-time synthesis.
In summary, we make the following contributions:
\begin{itemize}
\item We introduce \methodname, a dynamic scene representation that couples 3D Gaussians with neural ODEs to model the continuous, non-rigid evolution of plant growth from multi-view time-lapse images.
\item We propose a reverse-growth formulation that sidesteps non-differentiable topology changes and enables end-to-end training of a continuously evolving scene representation.
\item To the best of our knowledge, we present the first multi-view timelapse dataset of real growing plants, comprising three plant species (\textit{blooming flower}, \textit{corn}, and \textit{paperwhite}) recorded using a calibrated single-camera turntable system.
\end{itemize}

\paragraph{Opportunities for future research.}
Modeling the dynamic topological changes associated with plant growth is a challenging problem, but research in this direction has strong potential for scientific impact. 
We therefore overview the limitations of our current formulation alongside the many opportunities it opens for future research.
First, \methodname is optimized for monotonic growth scenarios.
While we show that the approach performs well on real captured data, extending it to processes involving structural loss, such as leaf senescence or petal drop, is a natural and promising direction.
Second, while this work focuses on temporally coherent geometry reconstruction and novel-view synthesis, coupling our representation with explicit trait extraction modules could unlock direct recovery of morphogenetic quantities—such as stem length, leaf area, and branching angles—opening exciting new avenues for automated plant phenotyping and monitoring. To facilitate future work, we will publicly release all code and data.

\section{Related Work}
\label{sec:related_works}

\vspace{-0.5em}
\paragraph{Dynamic novel view synthesis.}
Recent work in dynamic 3D scene modeling has largely shifted from Neural Radiance Fields (NeRFs)~\cite{mildenhall2021nerf, park2021nerfies} to 4D extensions of 3D Gaussian Splatting (3DGS)~\cite{kerbl20233d}, which offer superior rendering quality and computational efficiency.
The most common strategy is to learn a deformation field that maps a single set of canonical Gaussians to their state at each observed timestep~\cite{wu20244d, yang2024deformable, duisterhof2023deformgs, huang2024sc, liu2025dynamic}.
This process is often accelerated using compact and efficient neural representations such as HexPlanes~\cite{cao2023hexplane,fridovich2023k}.
However, deformation-based representations learn independent per-timestep deformations from a canonical space; as a result, they do not explicitly introduce new structure or capture the local spatio-temporal dependencies and monotonic growth inherent in plant growth.

Another line of work optimizes 4D spatio-temporal Gaussians to represent the scene’s evolution~\cite{yang20244d, duan20244d, li2024spacetime}.
A related approach models the continuous trajectory of each Gaussian’s parameters over time, often using simple functions such as polynomials~\cite{lin2024gaussian, wang2024shape}.
Finally, some methods adopt a sequential strategy, propagating Gaussian parameters from one frame to the next to enforce temporal consistency~\cite{luiten2024dynamic}.
However, these methods often rely on auxiliary inputs (e.g., optical flow or depth) or use masks to remove "inactive" Gaussians, which breaks explicit 3D correspondences between timesteps; sequential methods additionally assume persistent structures, and cannot account for new geometry emerging over time. In contrast, our approach models plant growth as a continuous, temporally coherent 3D Gaussian flow, enabling both the introduction of new structures and accurate prediction of unseen timesteps.

\paragraph{Continuous-time dynamics models.}
Continuous-time dynamical systems can be mathematically represented as ordinary differential equations (ODEs), where the rate of change of the system state is described as a function of the current state and time. Neural ODEs~\cite{chen2018neural} parametrize the underlying flow field using a neural network and recover the trajectory of the system by integration. Several extensions focus on improving optimization stability~\cite{dupont2019augmented, finlay2020train}, computational efficiency~\cite{kelly2020learning, norcliffe2023faster, kidger2021hey}, or adapting them to irregularly sampled data~\cite{rubanova2019latent, goyal2022neural}.

Our work is most closely related to methods that model continuous-time dynamics of 3D scenes using neural ODEs.
For example, Du et al.~\cite{du2021neural} learn a velocity field by integrating an ODE over point tracks, but they require dense point correspondences as input. More recently, Wang et al.~\cite{wang2025ode} combined latent ODEs with 3D Gaussians for temporal forecasting; however, their primary goal is motion extrapolation beyond observed trajectories, whereas we introduce a new dynamic 3D Gaussian representation and a multi-stage optimization procedure specifically designed to capture plant growth.

While several prior techniques~\cite{zheng20174d, dong20174d, adebola2025growsplat, lobefaro2024spatio, pan2021multi, chebrolu2020spatio} tackle plant growth reconstruction, these methods rely on point cloud registration rather than modeling continuous-time dynamics with 3D Gaussians,  limiting their ability to interpolate between observations and to guarantee smooth trajectories, as our neural ODE representation does.

\section{Method}

Given a set of posed images ${\image_{\view}^{\timestep}}$ of a growing plant observed over multiple timesteps $t \in {0, \dots, \lastime}$ and multiple views $\view$, our goal is to reconstruct the plant’s growth in 3D such that the reconstruction faithfully follows its natural trajectory. In particular, we seek a representation that evolves smoothly over time while ensuring that the visible volume of the plant is monotonically non-decreasing, consistent with natural growth.
\begin{figure*}
    \includegraphics[width=\textwidth]{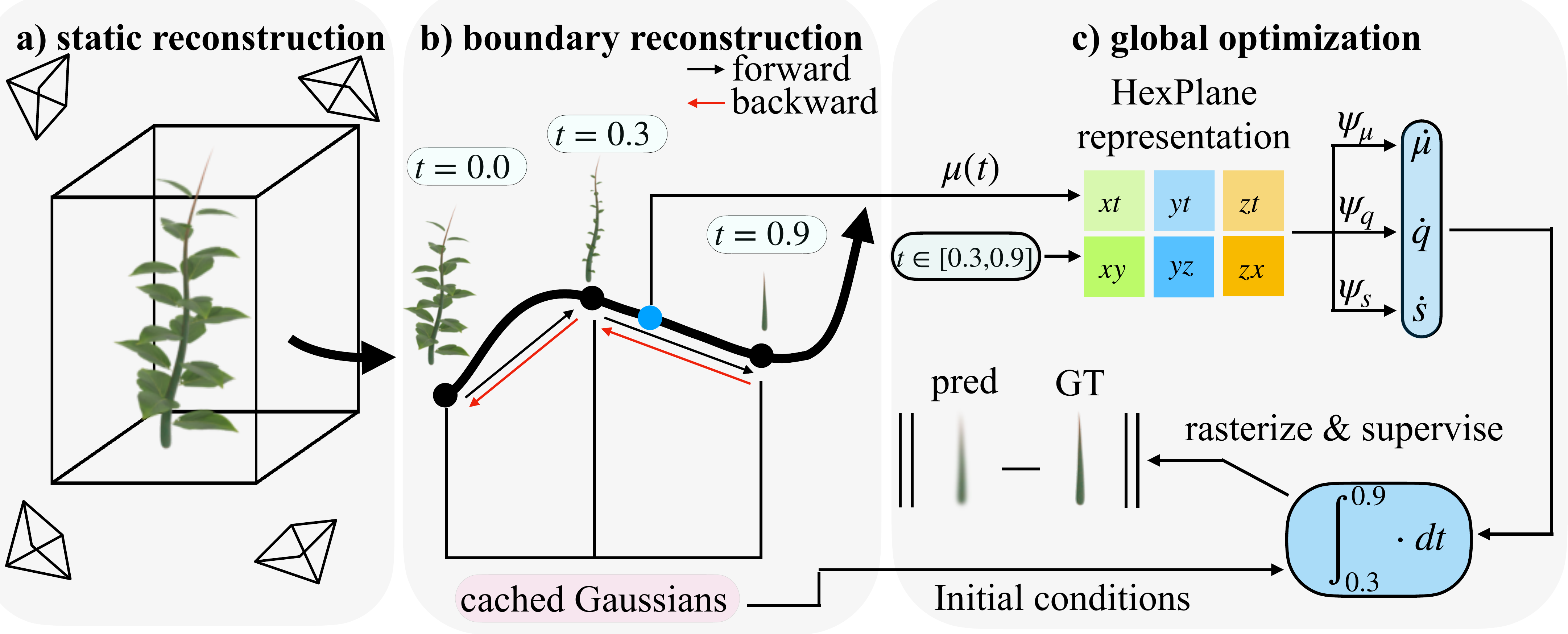}
    \caption{\textbf{Method overview.} \textbf{(a)} Our method first optimizes a set of 3D Gaussians on the fully-grown plant. \textbf{(b)} Using the optimized 3D Gaussians from the fully-grown plant, we progressively train the dynamics model to learn the state of the plant at each timestep. After each reconstructed timestep, we cache the Gaussians for that timestep and use them as initial conditions to optimize for the next timestep. \textbf{(c)} During the global optimization step, we randomly sample a timestep $t_k$ and integrate to $t_{k+1}$, leveraging the cached Gaussians from the boundary reconstruction step as initial conditions. We then optimize the dynamics model to enforce consistency between rendered and captured measurements.}
    \label{fig:method}
\end{figure*}
To this end, we adopt 3D Gaussian splats~\cite{kerbl20233d} as our underlying 3D representation and optimize a flow field that continuously evolves the Gaussian particles over time to model plant growth. Achieving such smooth temporal evolution is non-trivial: while existing approaches to dynamic 3D reconstruction allow arbitrary deformations either from a canonical template~\cite{wu20244d} or between discrete timesteps~\cite{luiten2024dynamic}, these formulations are not well-suited to modeling growth. Instead, plant growth should evolve continuously from one timestep to the next, following a smooth and monotonic trajectory rather than resetting from a canonical state or diverging unpredictably across timesteps.

To address this challenge, we first introduce a differentiable approach to modeling growth with 3D Gaussian particles in Section~\ref{sec:one}. We then develop a time-integrated neural field that produces a smooth trajectory of growth across all timesteps in Section~\ref{sec:two}. Finally, we present a training strategy that ensures stable optimization in Section~\ref{sec:three}.

\subsection{3D Gaussian Flow Fields}
\label{sec:one}
\vspace{-0.5em}
We represent the underlying 3D structure using 3D Gaussian Splatting (3DGS)~\cite{kerbl20233d}, a high-quality representation that enables real-time rendering. Specifically, the 3D scene is modeled with a set of $\numberg$ Gaussians ${\Gaussian_i}$, each parameterized by a center $\centr_i \in \mathbb{R}^3$, rotation quaternion $\quat_i \in \mathbb{R}^4$, scale $\scale_i \in \mathbb{R}^3$, opacity $\opacity_i \in \mathbb{R}$, and color coefficients $\colour_i \in \mathbb{R}^\rank$, represented via rank-$\rank$ spherical harmonics. These Gaussians are projected into a given view using a linearized projection model~\cite{zwicker2001ewa} and then alpha-blended in depth order to render the target image.

To model plant growth, we adapt this representation so that it evolves over time, allowing new structures to emerge gradually and coherently rather than being introduced abruptly. Growth can manifest in two ways: (i) increasing the scale of existing particles, thereby expanding the volume, or (ii) introducing new particles. While scale growth suffices at early stages, it cannot account for the formation of new matter and quickly degrades visual quality without particle addition. Conversely, densification in 3DGS is a discrete, non-differentiable process, making optimization challenging.
To address this, we reverse the problem: instead of modeling forward growth, we model backward shrinkage from the final state (time $t {=} T$) to the initial state ($t {=} 0$). This assumes that all matter required for the plant is already represented at $T$, eliminating the need for discrete particle addition. The task then reduces to making Gaussians disappear or “shrink” smoothly, either by scaling them down to zero or by becoming occluded within existing matter. This disappearance process is differentiable, making it well-suited for gradient-based optimization.
Consequently, the problem reduces to modeling the temporal deformation of Gaussian parameters that govern geometry while keeping appearance fixed. Concretely, we allow the center, rotation, and scale of each Gaussian 
to evolve over time, while assuming that color and opacity remain constant under fixed lighting conditions. This assumption is practical for our 
controlled capture setup, though the framework can naturally be extended to model time-varying appearance by including color in the flow field integration.
Each Gaussian is thus represented as
\begin{equation}
\Gaussian_i^{(t)} = \big(\centr_i^{(t)}, \quat_i^{(t)}, \scale_i^{(t)}, \opacity_i, \colour_i\big),
\end{equation}
where $\centr_i^{(t)}$, $\quat_i^{(t)}$, and $\scale_i^{(t)}$ are time-varying geometric parameters, and $\opacity_i$ and $\colour_i$ are time-invariant appearance parameters.

\subsection{Time-Integrated Velocity Field}
\label{sec:two}
\vspace{-0.5em}
Our goal is to obtain a smooth trajectory of growth by continuously deforming the geometry of Gaussians as they shrink backward in time. To this end, we model the velocities of Gaussian geometric parameters: translational velocity $\dot{\centr}_i(t)$, rotational velocity $\dot{\quat}_i(t)$, and volumetric velocity $\dot{\scale}_i(t)$. We define a time-dependent velocity field $\func$ governing the dynamics of each Gaussian:
\begin{equation}
\dot{\theta}_i(t) = \func(\centr_i(t), t),
\quad
\theta_i(t) = \theta_i(T) + \int_{T}^{t} \func(\centr_i(\tau), \tau) d\tau,
\end{equation}
where $\theta_i(t)$ denotes the geometric parameters of Gaussian $i$ at time $t$.
We require $\func$ to be at least $C^0$-continuous in both space and time. This guarantees that integrating the velocity field produces $C^1$-continuous trajectories, yielding smooth temporal evolution of centers, rotations, and scales. This design avoids sudden or unpredictable changes between timesteps, ensuring that the reconstructed plant evolves along smooth and differentiable trajectories.
We model the velocity field $\func$ using a spatio-temporal HexPlane encoder followed by multi-layer perceptron (MLP) decoders, similar to~\cite{wu20244d, cao2023hexplane}, as shown in Fig.~\ref{fig:method}. The HexPlane encoder interpolates features from a continuous spatio-temporal grid, which are then decoded by MLP heads into the geometric velocities.
Formally, given Gaussian centers $\centr_i(t)$ and time $t$, we extract a latent feature $\feat_i$ via:
\begin{equation}
\feat_i = \mlp\left(\text{HexInterp}(\centr_i(t), t)\right),
\end{equation}
where $\text{HexInterp}$ denotes interpolation from a multi-level HexPlane grid. Features are bilinearly interpolated from the six spatio-temporal planes $(x,y)$, $(y,z)$, $(x,z)$, $(x,t)$, $(y,t)$, $(z,t)$,  combined via a product across planes, and concatenated across $L$ resolution levels before being fed to the MLP $\mlp$.
The latent feature $\feat_i$ is then decoded into per-parameter velocities:
\begin{equation}
\dot{\centr}_i = \mlp_\mu(\feat_i),
\quad
\dot{\quat}_i = \mlp_q(\feat_i),
\quad
\dot{\scale}_i = \mlp_s(\feat_i),
\end{equation}
where $\mlp_\mu$, $\mlp_q$, and $\mlp_s$ are independent MLP decoders.
To recover Gaussian parameters at any future time $t_1$ from an initial state, we integrate velocity:
\begin{equation}
\theta_i(t_1) = \theta_i(t_0) + \int_{t_0}^{t_1} \func(\centr_i(t), t) dt,
\end{equation}
which can be solved using standard ODE solvers such as Runge–Kutta~\cite{butcher1996history,runge1895,kutta1901}.

\subsection{Training Dynamics}
\label{sec:three}
\vspace{-1em}
\paragraph{Static reconstruction.} 
We first optimize a static 3DGS model on the fully-grown plant at timestep $\lastime$, following standard procedure as in~\cite{kerbl20233d}, optimizing a mixture of L1 and SSIM losses. After optimization, we obtain a set of Gaussians $\Gaussian^{t_0} = \{\centr^{t_0}, \quat^{t_0}, \scale^{t_0}, \colour, \opacity\}$. 

\paragraph{Boundary reconstruction.} In principle, integrating from $t_0 = \lastime$ backward to all timesteps could produce the entire trajectory. However, directly optimizing such long-range ODE integration leads to unstable training, with vanishing gradients and accumulated numerical error. To address this, we adopt a piecewise integration strategy: instead of integrating across the full sequence, we train progressively from $\lastime$ to earlier steps $t_1, t_2, \dots$, caching intermediate states as boundary conditions. At each stage, the Gaussian state from the previous boundary condition $\Gaussian^{t_k}$ serves as the initial condition, and we integrate the velocity field through a single timestep to obtain $\Gaussian^{t_{k+1}}$: \begin{equation} \Gaussian^{t_{k+1}} = \Gaussian^{t_k} + \int_{t_k}^{t_{k+1}} \func(\centr(t), t)\, dt. \end{equation} This reduces the depth of recursive integration, stabilizes optimization, and ensures that each segment remains well-conditioned. Importantly, although integration is performed in a piecewise manner, the velocity field $F_\theta$ is shared across all segments, which guarantees continuity of the underlying dynamics. At each timestep, we supervise the predicted boundary state with an L1 loss against the ground-truth images of that timestep, and progressively expand the cache of boundary states as training proceeds. 

\paragraph{Global optimization.}
After recovering and storing all boundary states in the cache, we perform a global optimization of the trajectory. At each iteration, we randomly sample a timestep $t_k$ and integrate the velocity field between $t_k$ and $t_{k+1}$ using the cached boundary $\Gaussian^{t_k}$ as the initial condition: \begin{equation} \tilde{\Gaussian}^{t_{k+1}} = \Gaussian^{t_k} + \int_{t_k}^{t_{k+1}} \func(\centr(t), t)\, dt. \end{equation} The predicted Gaussians $\tilde{\Gaussian}^{t_{k+1}}$ are then rasterized and supervised against the ground truth images at timestep $t_{k+1}$ using an L1 penalty between the rendered and ground-truth pixel values.

\section{Multi-View Plant Growth Dataset}
\vspace{-1em}
\paragraph{Simulated dataset.}
We construct a simulated multiview timelapse dataset in Blender by porting seven distinct plant-growth scenes—\textit{clematis}, \textit{tulip}, \textit{plant1}, \textit{plant2}, \textit{plant3}, \textit{plant4}, and \textit{plant5}—originally created by artists on Blender Market. For each scene, we render 70 timesteps of growth from 34 camera viewpoints uniformly distributed along a full $360^\circ$ orbit around the plant, at a resolution of $400 \times 400$. This synthetic setup provides full control over geometry, materials, and lighting, enabling quantitative evaluation of reconstruction accuracy. For the spatial split, we use 31 views for reconstruction and 3 held-out views for novel-view evaluation at each timestep. For evaluation, we train on every 6th timestep (12 training timesteps, 372 training images per scene) and evaluate across 69 of 70 timesteps, of which 58 are unseen during training.

\begin{wrapfigure}[19]{r}{0.35\textwidth}
    \centering
    \vspace{-2.5em}
    \includegraphics[width=0.35\textwidth]{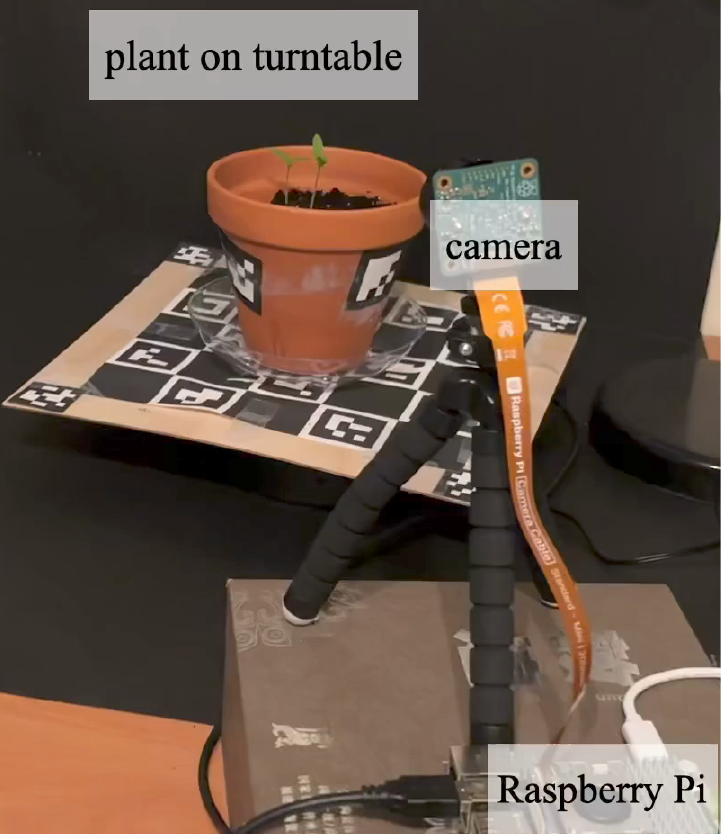}
    \caption{Multi-view timelapse capture setup. A Raspberry Pi-controlled turntable and camera autonomously capture multi-view images of the plant over multiple weeks.}
    \label{fig:hardware_figure}
\end{wrapfigure}
\paragraph{Captured dataset.}
Our captured dataset consists of three plant scenes — \textit{blooming flower}, \textit{corn}, and \textit{paperwhite} — captured with a Raspberry Pi HQ camera~\cite{upton2016raspberry} (Fig.~\ref{fig:hardware_figure}). The three species were chosen to represent a diverse range of growth patterns and temporal scales, with each sequence focused on the most dynamic phase of development: the \textit{blooming flower} undergoes rapid petal expansion, \textit{corn} exhibits strong vertical elongation and leaf splitting, and \textit{paperwhite} displays complex branching with multiple structures emerging simultaneously. Plants are placed on a motorized turntable; at each timestep, we capture 50 images at fixed elevation with 7.2° angular spacing, yielding full 360° coverage. We use 43 views for reconstruction and 7 held-out views for novel-view evaluation at each timestep. Images are captured at a resolution of $1200 \times 1200$. Capture frequency is adapted to each species' growth rate: for the \textit{blooming flower}, we capture every 15 minutes for 86 timesteps (4,300 total images); for \textit{corn}, every hour for 71 timesteps (3,550 total images); and for \textit{paperwhite}, every hour for 50 timesteps (2,500 total images). For evaluation, we train on a sparse subset of timesteps and evaluate across the full sequence. For \textit{blooming flower}, \textit{corn}, and \textit{paperwhite}, we train on every 17th, 10th, and 7th timestep respectively (6, 8, and 8 training timesteps; 258, 344, and 344 training images), evaluating on all 86, 71, and 50 timesteps, of which 80, 63, and 42 are unseen.

To get poses for training, we run COLMAP \cite{schonberger2016structure} on all images of the first timestep and propagate them to the other timesteps as the viewpoints are the same throughout.

\begin{figure*}[!ht]
    \centering
    \includegraphics[width=\textwidth]{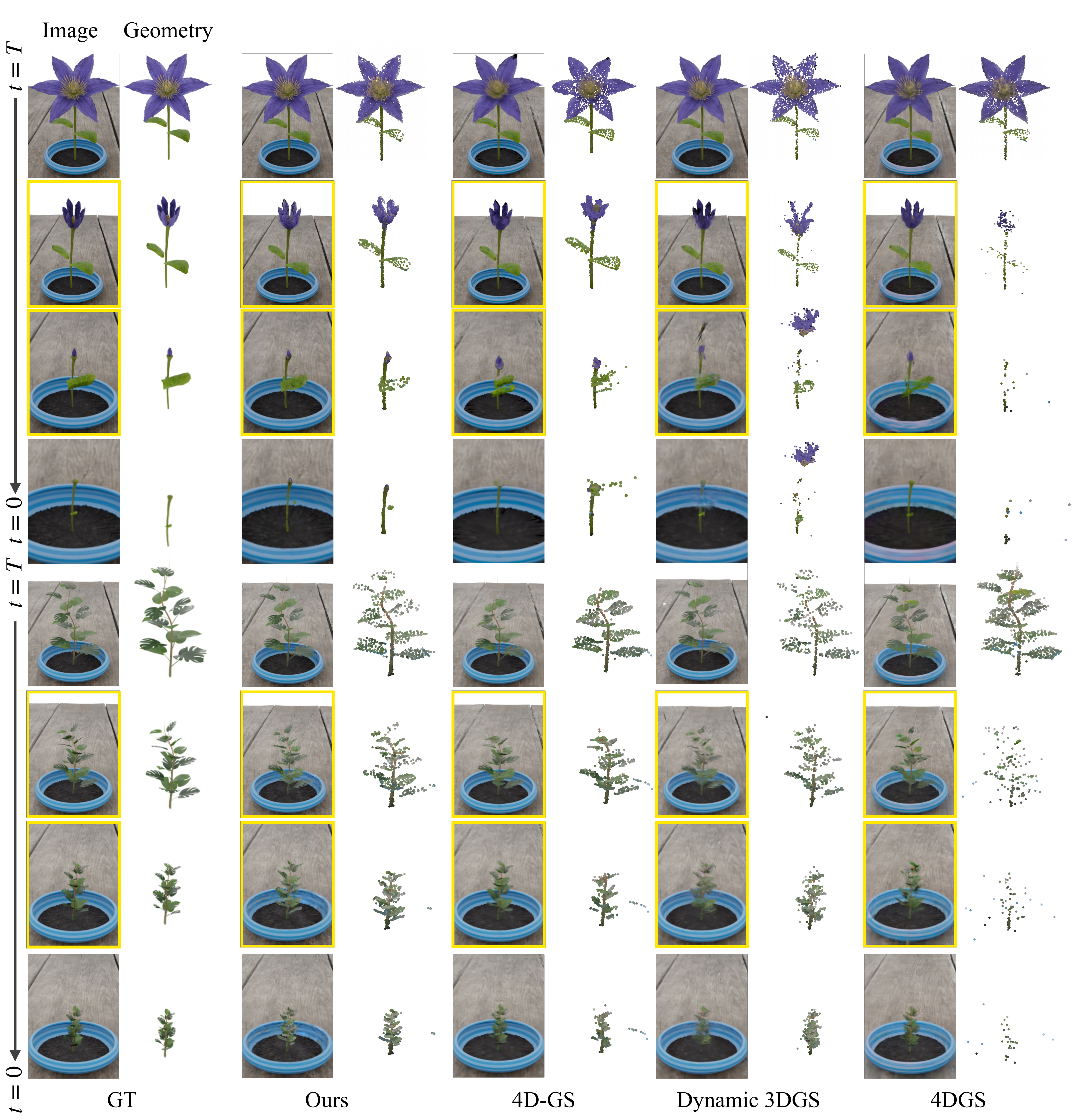}
\setlength{\tabcolsep}{5pt}
    \resizebox{\textwidth}{!}{
\begin{tabular}{l|cccc|cccc|cccc}
\toprule
        & \multicolumn{4}{c|}{Training times} & \multicolumn{4}{c|}{Interpolation times} & \multicolumn{4}{c}{Combined} \\
        \textbf{Method} & PSNR (dB)$\,\uparrow$ & SSIM$\,\uparrow$ & LPIPS$\,\downarrow$ & CD$\,\downarrow$ & PSNR (dB)$\,\uparrow$ & SSIM$\,\uparrow$ & LPIPS$\,\downarrow$ & CD$\,\downarrow$ & PSNR (dB)$\,\uparrow$ & SSIM$\,\uparrow$ & LPIPS$\,\downarrow$ & CD$\,\downarrow$ \\\midrule
        4D-GS & \cellcolor{tabsecond}{33.04} & \cellcolor{tabsecond}{0.946} & \cellcolor{tabsecond}{0.094} & \cellcolor{tabsecond}{0.73} & \cellcolor{tabsecond}{32.77} & \cellcolor{tabsecond}{0.944} & \cellcolor{tabsecond}{0.094} & \cellcolor{tabsecond}{0.78} & \cellcolor{tabsecond}{32.81} & \cellcolor{tabsecond}{0.944} & \cellcolor{tabsecond}{0.094} & \cellcolor{tabsecond}{0.77} \\
        4DGS & {30.19} & \cellcolor{tabthird}{0.939} & \cellcolor{tabthird}{0.107} & \cellcolor{tabthird}{12.00} & {29.11} & {0.905} & \cellcolor{tabthird}{0.145} & \cellcolor{tabthird}{11.95} & {29.29} & \cellcolor{tabthird}{0.910} & \cellcolor{tabthird}{0.138} & \cellcolor{tabthird}{11.96} \\
        Dynamic 3DGS & \cellcolor{tabthird}{32.48} & {0.912} & {0.154} & {13.18} & \cellcolor{tabthird}{32.03} & \cellcolor{tabthird}{0.908} & {0.158} & {13.64} & \cellcolor{tabthird}{32.11} & {0.909} & {0.157} & {13.56} \\
        Ours & \cellcolor{tabfirst}{35.43} & \cellcolor{tabfirst}{0.957} & \cellcolor{tabfirst}{0.065} & \cellcolor{tabfirst}{0.10} & \cellcolor{tabfirst}{34.93} & \cellcolor{tabfirst}{0.955} & \cellcolor{tabfirst}{0.066} & \cellcolor{tabfirst}{0.11} & \cellcolor{tabfirst}{35.02} & \cellcolor{tabfirst}{0.956} & \cellcolor{tabfirst}{0.066} & \cellcolor{tabfirst}{0.11} \\
        \bottomrule

    \end{tabular}
}
    \caption{
    \textbf{Results on synthetic data.} We compare results on both seen and interpolated times averaged over synthetic scenes. {GrowFlow} achieves stable geometry, unlike prior methods that show visually correct renderings for training frames but struggle on interpolation frames. \textcolor{darkyellow}{Yellow} marks interpolated frames, and $\downarrow$ next to a metric indicates that a lower value is better. Please see the Supp.\ Webpage for video results.
    }
    \vspace{-10px}
    \label{fig:synthetic_figure}
\end{figure*}
\begin{figure*}[!ht]
    \centering
    \includegraphics[width=\textwidth]{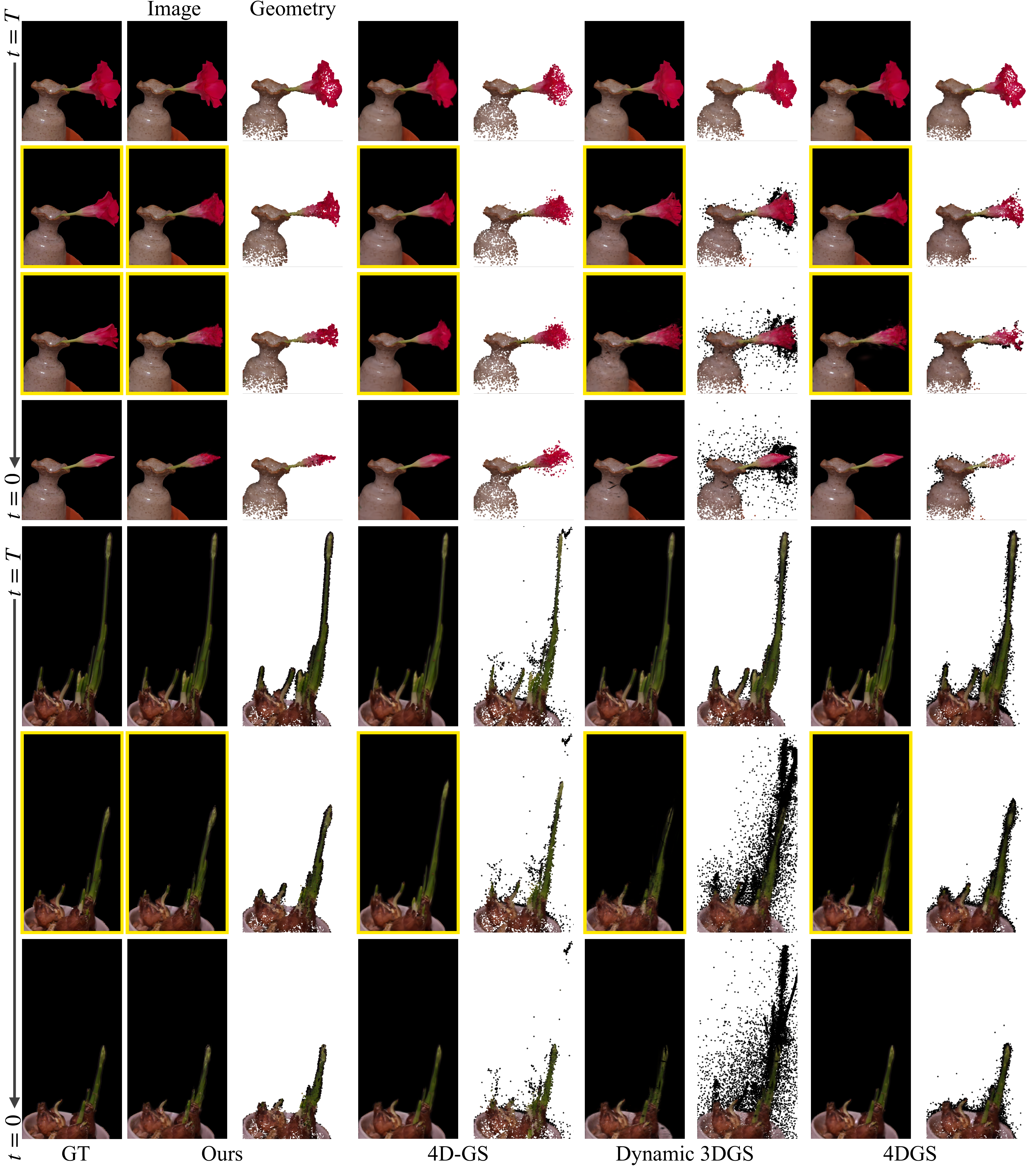}
\setlength{\tabcolsep}{5pt}
    \resizebox{\textwidth}{!}{
\begin{tabular}{l|ccc|ccc|ccc}
        \toprule
        & \multicolumn{3}{c|}{Training times} & \multicolumn{3}{c|}{Interpolation times} & \multicolumn{3}{c}{Combined} \\
        \textbf{Method} & PSNR (dB)$\,\uparrow$ & SSIM$\,\uparrow$ & LPIPS$\,\downarrow$ & PSNR (dB)$\,\uparrow$ & SSIM$\,\uparrow$ & LPIPS$\,\downarrow$ & PSNR (dB)$\,\uparrow$ & SSIM$\,\uparrow$ & LPIPS$\,\downarrow$ \\\midrule
        4D-GS & \cellcolor{tabsecond}{30.51} & \cellcolor{tabsecond}{0.989} & \cellcolor{tabsecond}{0.036} & \cellcolor{tabthird}{24.70} & \cellcolor{tabthird}{0.977} & \cellcolor{tabsecond}{0.045} & \cellcolor{tabthird}{25.36} & \cellcolor{tabthird}{0.978} & \cellcolor{tabsecond}{0.044} \\
        4DGS & \cellcolor{tabfirst}{31.25} & \cellcolor{tabfirst}{0.991} & \cellcolor{tabfirst}{0.031} & \cellcolor{tabsecond}{26.86} & \cellcolor{tabsecond}{0.979} & \cellcolor{tabthird}{0.046} & \cellcolor{tabsecond}{27.37} & \cellcolor{tabsecond}{0.981} & \cellcolor{tabsecond}{0.044} \\
        Dynamic 3DGS & {27.49} & {0.981} & \cellcolor{tabthird}{0.049} & {23.86} & {0.960} & {0.075} & {24.27} & {0.963} & \cellcolor{tabthird}{0.072} \\
        Ours & \cellcolor{tabthird}{28.80} & \cellcolor{tabthird}{0.987} & \cellcolor{tabfirst}{0.031} & \cellcolor{tabfirst}{27.28} & \cellcolor{tabfirst}{0.984} & \cellcolor{tabfirst}{0.033} & \cellcolor{tabfirst}{27.47} & \cellcolor{tabfirst}{0.984} & \cellcolor{tabfirst}{0.033} \\
        \bottomrule
    \end{tabular}
}

    \caption{
    \textbf{Results on captured data.} We compare results on both seen (``training'') and interpolated times averaged over all captured scenes. {GrowFlow} achieves stable, coherent geometry, unlike prior methods that struggle with renderings and reconstructed geometry on the interpolated frames. \textcolor{darkyellow}{Yellow} marks interpolated frames, and $\downarrow$ next to a metric indicates that a lower value is better. Please see the Supp.\ Webpage for video results.
    }

    \vspace{-10px}
    \label{fig:real_figure}
\end{figure*}
\begin{figure*}
    \begin{center}
    \includegraphics[width=\linewidth]{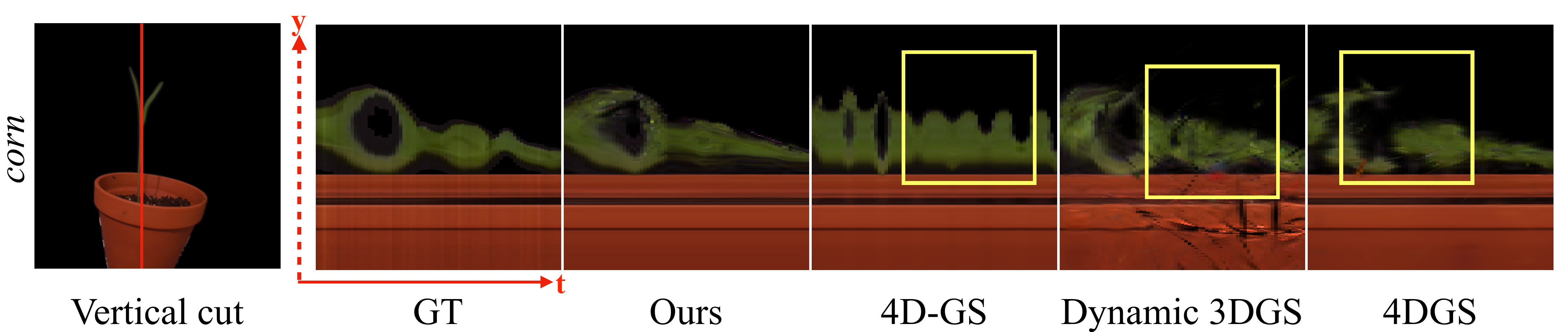}
    \end{center}
    \vspace{-6px}
    \caption{\textbf{Temporal slice visualization.} We analyze the accuracy of reconstructed motion by tracking a vertical cut from the predicted images of the \textit{corn} scene through time. Our method shows more faithful alignment with GT, while baselines exhibit noisy temporal dynamics (\textcolor{darkyellow}{yellow} boxes).}
    \label{fig:cut_figure}
\end{figure*}
\section{Experiments}
\label{sec:results}


\paragraph{Implementation details.}
For static reconstructions of fully grown plants, we use 3DGS with default training settings and the Adam~\cite{kingma2014adam} optimizer, training each model for 30K iterations. During the boundary reconstruction phase, we optimize each boundary timestep for 300 iterations using the adjoint method~\cite{chen2018neural}, with relative and absolute tolerances of $10^{-4}$ and $10^{-5}$, respectively, for the neural ODE solver. The dynamic reconstruction phase uses the same solver configuration and is trained for 30K iterations. 

\paragraph{Baselines.}
We compare our method against state-of-the-art methods in dynamic reconstruction: Dynamic 3DGS~\cite{luiten2024dynamic}, 4D-GS~\cite{wu20244d}, and 4DGS ~\cite{yang20244d}. For all results, we use the corresponding open source implementations of these methods.
For timestep interpolation, our method, 4D-GS, and 4DGS inherently support querying intermediate timesteps. For Dynamic 3DGS, which does not natively support continuous time, we perform interpolation between learned timesteps by fitting a third-degree polynomial to the Gaussian centers and colors. Rotations are interpolated using spherical linear interpolation (slerp), while scales and opacities are kept fixed, consistent with the original implementation.

\paragraph{Metrics.}
We employ two complementary measures to evaluate reconstruction methods. Since our goal is to recover geometrically faithful growth rather than only achieving photometric accuracy, we introduce a geometric accuracy metric based on Chamfer Distance (CD). 
We track foreground Gaussians by matching each to its nearest vertex on the ground-truth plant mesh at the first timestep. Per-timestep Chamfer Distance is then computed between these foreground Gaussians and their corresponding mesh vertices, averaged across time. For 4DGS, we apply their temporal masking before computing distances.
In addition, we evaluate the photometric quality of test views using standard image-based metrics: PSNR, LPIPS, and SSIM.

\vspace{-1em}
\subsection{Simulated Results}
\paragraph{Qualitative comparisons.}
\cref{fig:synthetic_figure} presents qualitative and quantitative comparisons against baseline methods for plant-growth reconstruction. Our method yields geometrically coherent trajectories: Gaussian centers closely follow the plant’s true surface over time and produce high-quality novel-view renderings. In contrast, baseline approaches exhibit pronounced geometric drift, with Gaussian centers gradually detaching from the plant surface or floating in space as time progresses. Dynamic 3DGS \cite{luiten2024dynamic} and 4D-GS \cite{wu20244d} frequently displace Gaussians corresponding to shrunken or disappearing structures into the far field or behind background elements, rather than shrinking them downward as the plant regresses. As illustrated in \cref{fig:synthetic_figure}, these Gaussians often remain at roughly their original height but are simply pushed behind the scene, making them invisible in the renderings. Furthermore, 4DGS \cite{yang20244d} leverages different Gaussians to model different frames separately, limiting its ability to track the same set of Gaussians throughout time. 

These behaviors highlight a key limitation of approaches that do not explicitly model continuous growth: they prioritize reproducing photorealistic appearance in training views at the expense of temporally coherent geometry. Our representation optimizes a smooth flow field over Gaussian parameters, allowing superior novel view synthesis capabilities, but most importantly, reconstructing physically plausible growth.

\paragraph{Quantitative comparisons.}
Quantitatively, our approach outperforms all baselines by a substantial margin in both image-quality metrics and Chamfer Distance. This demonstrates that \methodname achieves superior geometric fidelity and photometric consistency not only at supervised training timesteps but also at the 58 interpolated timesteps unseen during training.

\subsection{Captured Results}
\paragraph{Qualitative comparisons.}
\cref{fig:real_figure} presents qualitative and quantitative comparisons against baseline methods on the \textit{blooming flower} and \textit{paperwhite} scenes. While baselines render novel views at training timesteps well, their quality degrades when rendering novel views at interpolated timesteps. 4D-GS~\cite{wu20244d} fails most notably during interpolation: rather than producing smooth shrinkage, the reconstructed plant oscillates between growing and shrinking. Dynamic 3DGS~\cite{luiten2024dynamic} assumes fixed Gaussian sizes over time and thus cannot model the shrinking plant; it instead turns affected Gaussians black to match the background, minimizing photometric loss at the cost of physical plausibility. In contrast, our method produces temporally smooth and physically plausible interpolations throughout.

\paragraph{Quantitative comparisons.}
We omit the Chamfer Distance calculation as we do not have ground-truth mesh for the captured data. Overall, our method achieves higher quality novel view renderings compared to baseline methods. Despite achieving slightly lower PSNR and SSIM on the training timesteps, our LPIPS is comparable to baselines. Because our neural ODE optimizes for a continuous flow field of Gaussian parameters rather than overfitting to individual training timesteps, it trades slightly lower performance on training timesteps for superior interpolation quality on real-world plants. Nonetheless, it produces more plausible growth geometry versus baselines.

\paragraph{Temporal slice visualization.}
To further evaluate motion accuracy, \cref{fig:cut_figure} visualizes a tracked horizontal slice of the plant across timesteps in a novel rendered viewpoint for the \textit{corn} scene. Our method closely matches the ground-truth motion, whereas baselines exhibit significant structural distortions and temporal misalignment.

\subsection{Ablation Study}
\label{sec:ablation}

\begin{figure*}
    \begin{center}
    \includegraphics[width=\textwidth]{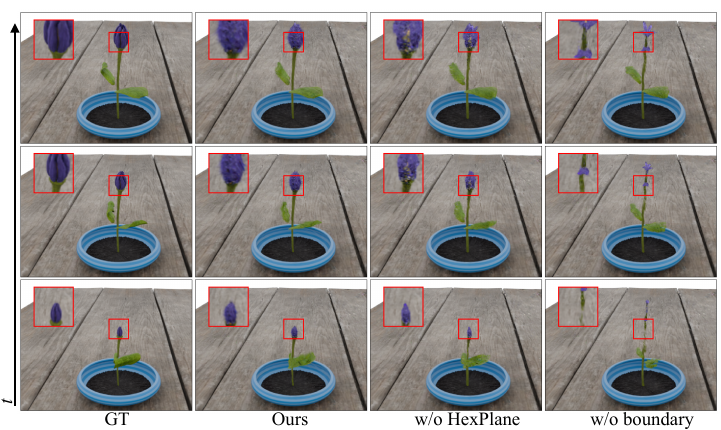}
    \end{center}
    \caption{\textbf{Qualitative ablations.} Replacing our HexPlane representation with an MLP with Fourier encodings reduces capacity and degrades rendering quality. Skipping the boundary reconstruction stage causes the reconstructed geometry to break down.}
    \vspace{-15px}
    \label{fig:ablation_figure}
\end{figure*}

\setlength{\intextsep}{5pt}
\begin{wraptable}[8]{r}{0.55\textwidth}
\centering
\captionof{table}{Ablation on the \textit{clematis} scene.}
\footnotesize
\begin{tabular}{lcccc}
\toprule
Method & PSNR~$\uparrow$ & SSIM~$\uparrow$ & LPIPS~$\downarrow$ & CD~$\downarrow$ \\
\midrule
Ours         & \cellcolor{tabfirst}{33.05} & \cellcolor{tabfirst}{0.947} & \cellcolor{tabfirst}{0.071} & \cellcolor{tabfirst}{0.02} \\
\midrule
w/o HexPlane & \cellcolor{tabsecond}{32.18} & \cellcolor{tabsecond}{0.944} & \cellcolor{tabsecond}{0.076} & \cellcolor{tabsecond}{0.03} \\
w/o boundary  & \cellcolor{tabthird}{28.52} & \cellcolor{tabthird}{0.914} & \cellcolor{tabthird}{0.097} & \cellcolor{tabthird}{36.47} \\
\bottomrule
\end{tabular}
\label{tab:ablation}
\end{wraptable}
\textbf{HexPlane.}
Neural ODE frameworks are often parameterized using MLPs. However, as shown in the insets of Fig.~\ref{fig:ablation_figure}, substituting our spatio-temporal HexPlane encoder with an MLP leads to noticeably degraded reconstruction quality, e.g., the flower bud exhibits more artifacts and temporal instability. HexPlane provides a higher-quality inductive bias for capturing spatial and temporal variations, enabling smoother and more consistent Gaussian trajectories. The quantitative results in Tab.~\ref{tab:ablation} confirm this, i.e., the HexPlane achieves superior image fidelity and improves geometric accuracy compared to the MLP alternative.

\paragraph{Boundary reconstruction.}
The boundary reconstruction stage is essential for stable optimization of the neural ODE. Without it, the model must rely on long-range integration from the final timestep to all earlier states, which leads to accumulated numerical errors, vanishing gradients, and poor convergence. Although the model can eventually produce reasonable photometric reconstructions, it struggles to maintain geometric consistency, resulting in drifting Gaussians and degraded temporal coherence. As shown in Fig.~\ref{fig:ablation_figure} and Tab.~\ref{tab:ablation}, removing the boundary reconstruction step substantially harms both image quality and geometric fidelity, highlighting its importance in accurately modeling continuous plant growth.

\vspace{-1em}
\section{Conclusion}
\label{sec:conclusion}
\vspace{-1em}
In this work, we propose \methodname, the first continuous dynamic 3D 
representation for plant growth, combining 3D Gaussians with neural ODEs 
to model the non-rigid evolution of plant growth from multi-view time-lapse 
images. By learning a continuous 3D Gaussian flow field, \methodname 
captures the underlying growth vector field, enabling temporally coherent 
reconstruction of plant geometry. To address the challenge of continuously 
emerging structures, we introduce a reverse-growth formulation, training 
the model to shrink 3D Gaussians over time and later reversing this flow 
to recover realistic growth trajectories. We validate our method on both 
synthetic scenes and a real-world captured dataset of three plant species 
--- \textit{blooming flower}, \textit{corn}, and \textit{paperwhite} --- 
recorded with a calibrated single-camera turntable system, demonstrating 
superior geometric accuracy and photometric quality compared to existing baselines.

\methodname is designed under the assumption of monotonic growth, which is directly relevant to many plant phenotyping and agricultural applications, and is practical for species exhibiting predominantly additive growth. We view this 
as a natural starting point for this problem, and encourage future work 
to relax this assumption to handle non-monotonic phenomena such as leaf 
senescence and pruning. Other promising directions include incorporating 
biologically motivated priors and extending the framework to other dynamic 
objects whose geometry emerges over time, e.g., growing crystals, 
developing embryos, or erupting geological formations.
\clearpage
\paragraph{Acknowledgements.}
DBL acknowledges support of NSERC under the RGPIN program. DBL also acknowledges support from the Canada Foundation for Innovation and the Ontario Research Fund.

\bibliographystyle{splncs04}
\bibliography{main}
\newpage
\appendix
\renewcommand{\thetable}{S\arabic{table}}
\setcounter{table}{0}
\renewcommand{\thefigure}{S\arabic{figure}}
\setcounter{figure}{0}
\begin{center}
    {\Large\textbf{Supplementary Material}}
\end{center}
\section{Video Results}
We include an extensive set of results in the \href{https://weihanluo.ca/growflow/}{Supp.\ Webpage}. There, we show novel view and geometry comparisons against baseline methods on synthetic and captured data. We further show the produced flow field from our trained model.
\section{Implementation Details}

In this section, we provide a
detailed description of the network architecture. We implement our dynamic Gaussian representation using the open-sourced Gaussian Splatting implementation gsplat \cite{ye2025gsplat} and the neural ODE codebase torchdiffeq \cite{chen14torchdiffeq}. Our HexPlane architecture follows closely \cite{wu20244d, cao2023hexplane}, where the spatial resolutions are set to 64 and the temporal resolution is set to 25, which are upsampled by 2. The learning rate of the HexPlane is set to $1.6\times 10^{-3}$, and the learning rate of the MLP decoder is set to $1.6\times 10^{-4}$, both of which are exponentially decayed by a factor of 0.1 until the end of training, for 30K iterations.  Unlike \cite{wu20244d}, we omit the total variation loss, as it does not bring additional improvement. We use a batch size of 30 viewpoints for both our boundary reconstruction stage and dynamic optimization stage, but keep the temporal batch size to 1. The MLP decoders consist of a two-layer MLP with 64 units and a ReLU activation function. 

After static reconstruction, we fix the background Gaussians and optimize only the foreground Gaussians within a manually defined bounding box. This constrains the neural ODE to modeling foreground flow, greatly easing optimization.

\section{Dataset}
Extra details of all the simulated and captured datasets can be found in Table \ref{tab:synth_dataset} and Table \ref{tab:cap_dataset}.

\paragraph{Hardware.} Our setup consists of a Raspberry Pi 5 (16GB) with an Active Cooler, powered by a 27W USB-C supply, and an HQ Camera CS with a 6mm wide-angle lens connected via a 300mm cable and stabilized on a tripod. The Pi sends commands to a programmable motorized turntable (ComXim) to rotate the plant and triggers the camera to capture images at each position. To prevent plants wobbling during captures, we set the velocity of the turntable to be the lowest and wait a few seconds after rotations before doing a capture.
A pseudo-code of the capture process is illustrated \ref{alg:data_collection_algo}.

\begin{algorithm}
\caption{Real Data Collection for \methodname}
\label{alg:data_collection_algo}
\begin{algorithmic}[1]
    \State \textbf{Hardware components:} Raspberry Pi, HQ Camera CS, motorized turntable. 
    \State Set turntable velocity to be lowest.
    \For{$t = 1$ to $n_{\text{timesteps}}$}
        \For{$p = 1$ to $n_{\text{views}}$}
            \State Send rotation command to turntable: rotate by $\frac{360}{n_{\text{views}}}$ degrees
            \State Wait for turntable to stabilize
            \State Trigger camera to capture image $I^{t}_{p}$
        \EndFor
        \State Wait until next timestep
    \EndFor
    
    \State \textbf{Output:} Multi-view image set $\{I^{t}_{p}\}$ for all timesteps $t$ and views $p$

\end{algorithmic}
\end{algorithm}

\begin{table}[h]
\centering
\caption{Descriptions of simulated scenes. All scenes sit in a blue vase on top of a wooden table.}
\label{tab:synth_dataset}
\begin{tabular}{|p{1.5cm}|p{10cm}|}
\hline
 & \multicolumn{1}{c|}{Scene description} \\
\hline
Clematis &  A purple clematis flower with six-pointed petals and yellow-white stamens at the center, growing on a thin green stem with small leaves. \\
\hline
Tulip & A pink tulip with partially open petals, showing a lighter pink/white interior, on a green stem with two long tulip-shaped leaves. \\
\hline
Plant1 & A small, green, young seedling with thin stems and small jagged leaves. \\
\hline
Plant2 & A young plant with a tall, dark central stem and several branches bearing smooth, rounded green leaves at varying heights. \\
\hline
Plant3 & A tall, slender plant with a pink-red stem and narrow, elongated leaves in shades of pink and purple, arranged sparsely along the stem. \\
\hline
Plant4 & A young, slender seedling with a single upright stem and several pairs of bright green, oval leaves arranged oppositely along the stalk. \\
\hline
Plant5 & A small plant with distinctively split and fenestrated dark green leaves, resembling a juvenile Monstera deliciosa, with multiple leaves branching from a central stem.\\
\hline
\end{tabular}
\end{table}

\begin{table}[h]
\centering
\caption{Descriptions of captured scenes. The growth time refers to the total duration from planting to the end of the capture period.}
\label{tab:cap_dataset}
\begin{tabular}{|p{1.7cm}|p{8cm}|p{2cm}|}
\hline
 & \multicolumn{1}{c|}{Scene description} & \multicolumn{1}{c|}{Growth time} \\
\hline
Paperwhite & A paperwhite narcissus with several brown bulbs sitting in a shallow white pot filled with gravel. Multiple green shoots are emerging from the bulbs, with one tall, slender stem extending prominently upward, topped with a closed bud. & one month\\
\hline
Corn & A corn seedling with two narrow, upright green leaves forming a V-shape, planted in a terracotta pot filled with dark soil. & three weeks\\
\hline
Blooming flower &A vibrant pink-red flower in full bloom with wide, open petals, growing from a bulbous beige ceramic vase-shaped pot sitting on an orange saucer.& one week \\
\hline
\end{tabular}
\end{table}
\section{Additional Results}
\subsection{Synthetic Results}

Tables \ref{tab:cross_scene_psnr_db}, \ref{tab:cross_scene_ssim}, \ref{tab:cross_scene_lpips}, \ref{tab:cross_scene_cd} provide a breakdown of the quantitative results in simulation across all scenes. Overall, our method achieves state-of-the-art performance across all scenes compared to baselines. Please refer to the Supp.\ Webpage for additional video results and comparisons to baselines.

\begin{table*}
    \caption{PSNR (dB) results across different synthetic scenes for combined (training + interpolation) frames.}
    \label{tab:cross_scene_psnr_db}
    \centering
    \begin{tabular}{lcccccccc}
        \toprule
        \textbf{Method} & Clematis & Plant1 & Plant2 & Plant3 & Plant4 & Plant5 & Tulip & Average \\
        \midrule
        4D-GS & \cellcolor{tabsecond}{31.10} & \cellcolor{tabsecond}{34.11} & \cellcolor{tabfirst}{33.11} & \cellcolor{tabsecond}{32.98} & \cellcolor{tabsecond}{34.30} & \cellcolor{tabsecond}{32.16} & \cellcolor{tabsecond}{31.94} & \cellcolor{tabsecond}{32.81} \\
        4DGS & 27.62 & 29.78 & 29.24 & 29.73 & 30.06 & 29.50 & 29.13 & 29.29 \\
        Dynamic 3DGS & \cellcolor{tabthird}{30.56} & \cellcolor{tabthird}{33.64} & \cellcolor{tabthird}{31.46} & \cellcolor{tabthird}{32.64} & \cellcolor{tabthird}{33.80} & \cellcolor{tabthird}{31.58} & \cellcolor{tabthird}{31.07} & \cellcolor{tabthird}{32.11} \\
        \midrule
        Proposed & \cellcolor{tabfirst}{33.05} & \cellcolor{tabfirst}{38.12} & \cellcolor{tabsecond}{32.73} & \cellcolor{tabfirst}{35.50} & \cellcolor{tabfirst}{37.54} & \cellcolor{tabfirst}{33.30} & \cellcolor{tabfirst}{34.90} & \cellcolor{tabfirst}{35.02} \\
        \bottomrule
    \end{tabular}
    \end{table*}

\begin{table*}
    \caption{SSIM results across different synthetic scenes for combined (training + interpolation) frames.}
    \label{tab:cross_scene_ssim}
    \centering
    \begin{tabular}{lcccccccc}
        \toprule
        \textbf{Method} & Clematis & Plant1 & Plant2 & Plant3 & Plant4 & Plant5 & Tulip & Average \\
        \midrule
        4D-GS & \cellcolor{tabsecond}{0.933} & \cellcolor{tabsecond}{0.952} & \cellcolor{tabfirst}{0.948} & \cellcolor{tabsecond}{0.946} & \cellcolor{tabsecond}{0.951} & \cellcolor{tabfirst}{0.942} & \cellcolor{tabsecond}{0.939} & \cellcolor{tabsecond}{0.944} \\
        4DGS & 0.887 & \cellcolor{tabthird}{0.922} & \cellcolor{tabthird}{0.911} & \cellcolor{tabthird}{0.914} & \cellcolor{tabthird}{0.921} & \cellcolor{tabthird}{0.910} & \cellcolor{tabthird}{0.908} & \cellcolor{tabthird}{0.910} \\
        Dynamic 3DGS & \cellcolor{tabthird}{0.900} & \cellcolor{tabthird}{0.922} & 0.903 & 0.913 & 0.920 & 0.905 & 0.901 & 0.909 \\
        \midrule
        Proposed & \cellcolor{tabfirst}{0.947} & \cellcolor{tabfirst}{0.968} & \cellcolor{tabsecond}{0.943} & \cellcolor{tabfirst}{0.963} & \cellcolor{tabfirst}{0.966} & \cellcolor{tabsecond}{0.941} & \cellcolor{tabfirst}{0.962} & \cellcolor{tabfirst}{0.956} \\
        \bottomrule
    \end{tabular}
    \end{table*}

\begin{table*}
    \caption{LPIPS results across different synthetic scenes for combined (training + interpolation) frames.}
    \label{tab:cross_scene_lpips}
    \centering
    \begin{tabular}{lcccccccc}
        \toprule
        \textbf{Method} & Clematis & Plant1 & Plant2 & Plant3 & Plant4 & Plant5 & Tulip & Average \\
        \midrule
        4D-GS & \cellcolor{tabsecond}{0.102} & \cellcolor{tabsecond}{0.087} & \cellcolor{tabsecond}{0.095} & \cellcolor{tabsecond}{0.095} & \cellcolor{tabsecond}{0.089} & \cellcolor{tabsecond}{0.097} & \cellcolor{tabsecond}{0.095} & \cellcolor{tabsecond}{0.094} \\
        4DGS & \cellcolor{tabthird}{0.158} & \cellcolor{tabthird}{0.129} & \cellcolor{tabthird}{0.139} & \cellcolor{tabthird}{0.136} & \cellcolor{tabthird}{0.130} & \cellcolor{tabthird}{0.140} & \cellcolor{tabthird}{0.135} & \cellcolor{tabthird}{0.138} \\
        Dynamic 3DGS & 0.162 & 0.148 & 0.165 & 0.156 & 0.152 & 0.161 & 0.155 & 0.157 \\
        \midrule
        Proposed & \cellcolor{tabfirst}{0.071} & \cellcolor{tabfirst}{0.051} & \cellcolor{tabfirst}{0.082} & \cellcolor{tabfirst}{0.061} & \cellcolor{tabfirst}{0.055} & \cellcolor{tabfirst}{0.089} & \cellcolor{tabfirst}{0.055} & \cellcolor{tabfirst}{0.066} \\
        \bottomrule
    \end{tabular}
    \end{table*}

\begin{table*}
    \caption{CD results across different synthetic scenes for combined (training + interpolation) frames.}
    \label{tab:cross_scene_cd}
    \centering
    \begin{tabular}{lcccccccc}
        \toprule
        \textbf{Method} & Clematis & Plant1 & Plant2 & Plant3 & Plant4 & Plant5 & Tulip & Average \\
        \midrule
        4D-GS & \cellcolor{tabsecond}{0.21} & \cellcolor{tabsecond}{0.20} & \cellcolor{tabsecond}{2.03} & \cellcolor{tabfirst}{0.22} & \cellcolor{tabsecond}{0.17} & \cellcolor{tabthird}{2.42} & \cellcolor{tabsecond}{0.12} & \cellcolor{tabsecond}{0.77} \\
        4DGS & \cellcolor{tabthird}{42.63} & 3.98 & 2.82 & 14.25 & 2.78 & 10.56 & \cellcolor{tabthird}{6.72} & \cellcolor{tabthird}{11.96} \\
        Dynamic 3DGS & 79.26 & \cellcolor{tabthird}{0.79} & \cellcolor{tabthird}{2.32} & \cellcolor{tabthird}{1.98} & \cellcolor{tabthird}{0.22} & \cellcolor{tabsecond}{0.40} & 9.98 & 13.56 \\
        \midrule
        Proposed & \cellcolor{tabfirst}{0.02} & \cellcolor{tabfirst}{0.08} & \cellcolor{tabfirst}{0.10} & \cellcolor{tabsecond}{0.28} & \cellcolor{tabfirst}{0.11} & \cellcolor{tabfirst}{0.12} & \cellcolor{tabfirst}{0.02} & \cellcolor{tabfirst}{0.11} \\
        \bottomrule
    \end{tabular}
    \end{table*}

\subsection{Captured Results}
\begin{figure*}[!ht]
    \begin{center}
    \includegraphics[width=\textwidth]{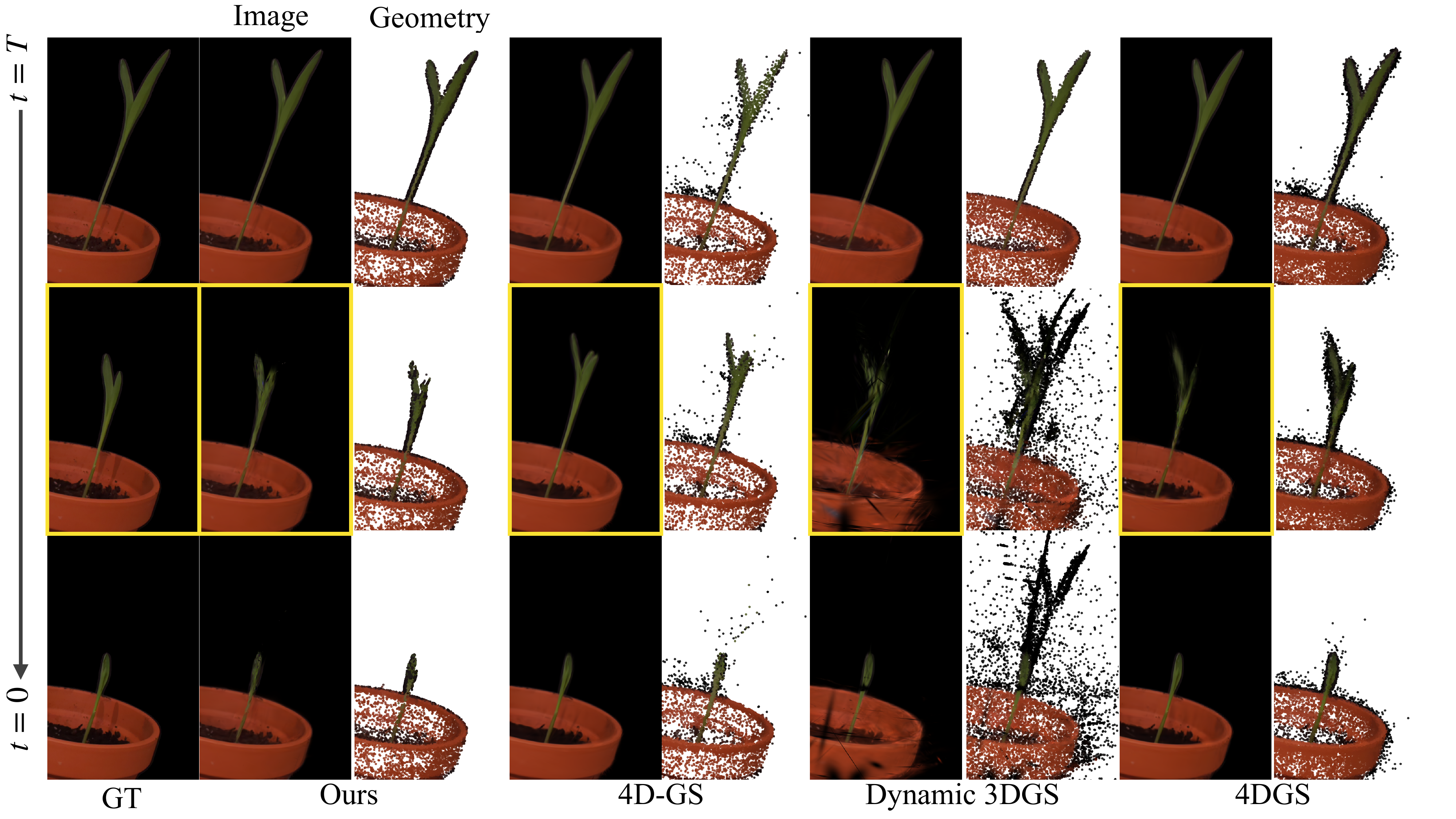}
    \end{center}
    \caption{We show our method's novel view renders against baselines on trained and interpolated timesteps. Our method more faithfully reconstructs the \textit{corn} at interpolated timesteps compared to baselines (images indicated with a yellow border are novel view renders of interpolated times).}
    \vspace{-10px}
    \label{fig:captured_figure_corn}
\end{figure*}

Tables \ref{tab:captured_cross_scene_psnr_db}, \ref{tab:captured_cross_scene_ssim}, \ref{tab:captured_cross_scene_lpips} provide a breakdown of the quantitative results across all captured scenes. Furthermore, Figure \ref{fig:captured_figure_corn} compares the reconstructed \textit{corn} scene across all baselines. Consistent with the results in the main text, our method reconstructs more accurate novel view renders and plant geometry over the training and interpolated timesteps. Please refer to the Supp.\ Webpage for additional video results and comparisons to baselines. 

\begin{table*}
    \caption{PSNR (dB) results across different captured scenes for combined (training + interpolation) frames.}
    \label{tab:captured_cross_scene_psnr_db}
    \centering
    \begin{tabular}{lcccc}
        \toprule
        \textbf{Method} & Blooming flower & Corn & Paperwhite & Average \\
        \midrule
        4D-GS & 23.11 & \cellcolor{tabthird}{29.22} & 23.74 & \cellcolor{tabthird}{25.36} \\
        4DGS & \cellcolor{tabfirst}{26.87} & \cellcolor{tabsecond}{30.12} & \cellcolor{tabfirst}{25.11} & \cellcolor{tabsecond}{27.37} \\
        Dynamic 3DGS & \cellcolor{tabthird}{25.26} & 23.46 & \cellcolor{tabthird}{24.09} & 24.27 \\
        \midrule
        Proposed & \cellcolor{tabsecond}{26.77} & \cellcolor{tabfirst}{30.68} & \cellcolor{tabsecond}{24.97} & \cellcolor{tabfirst}{27.47} \\
        \bottomrule
    \end{tabular}
    \end{table*}

\begin{table*}
    \caption{SSIM results across different captured scenes for combined (training + interpolation) frames.}
    \label{tab:captured_cross_scene_ssim}
    \centering
    \begin{tabular}{lcccc}
        \toprule
        \textbf{Method} & Blooming flower & Corn & Paperwhite & Average \\
        \midrule
        4D-GS & \cellcolor{tabthird}{0.983} & \cellcolor{tabthird}{0.982} & \cellcolor{tabthird}{0.970} & \cellcolor{tabthird}{0.978} \\
        4DGS & \cellcolor{tabsecond}{0.984} & \cellcolor{tabsecond}{0.984} & \cellcolor{tabsecond}{0.976} & \cellcolor{tabsecond}{0.981} \\
        Dynamic 3DGS & 0.976 & 0.946 & 0.966 & 0.963 \\
        \midrule
        Proposed & \cellcolor{tabfirst}{0.990} & \cellcolor{tabfirst}{0.986} & \cellcolor{tabfirst}{0.977} & \cellcolor{tabfirst}{0.984} \\
        \bottomrule
    \end{tabular}
    \end{table*}

\begin{table*}
    \caption{LPIPS results across different captured scenes for combined (training + interpolation) frames.}
    \label{tab:captured_cross_scene_lpips}
    \centering
    \begin{tabular}{lcccc}
        \toprule
        \textbf{Method} & Blooming flower & Corn & Paperwhite & Average \\
        \midrule
        4D-GS & \cellcolor{tabsecond}{0.032} & \cellcolor{tabsecond}{0.052} & \cellcolor{tabthird}{0.048} & \cellcolor{tabsecond}{0.044} \\
        4DGS & \cellcolor{tabthird}{0.034} & \cellcolor{tabthird}{0.053} & \cellcolor{tabsecond}{0.046} & \cellcolor{tabsecond}{0.044} \\
        Dynamic 3DGS & 0.044 & 0.111 & 0.062 & \cellcolor{tabthird}{0.072} \\
        \midrule
        Proposed & \cellcolor{tabfirst}{0.020} & \cellcolor{tabfirst}{0.043} & \cellcolor{tabfirst}{0.036} & \cellcolor{tabfirst}{0.033} \\
        \bottomrule
    \end{tabular}
\end{table*}

\section{\methodname Training Algorithm}
We begin with a detailed outline of the training algorithm of our pipeline in Algorithm \ref{alg:growflow training}.
The first phase is the static reconstruction stage, where we optimize a set of 3D Gaussians on posed images of the fully grown plant. By the end, we have optimized a set of Gaussians at timestep $t_0$, which we denote as $\Gaussian^{t_0}$. For the subsequent training phases, we freeze color $c$ and opacity $o$. Next, for the boundary reconstruction, we integrate backwards in time, one timestep at a time and cache the optimized Gaussians for each timestep. After this phase, we have a set of cached Gaussians for each timestep. Finally, during the global optimization step, we randomly sample a timestep, and leverage the cached Gaussian at that timestep to optimize the neural ODE. The result is a trained neural ODE $F_\phi$ able to interpolate over unseen timepoints.

\begin{algorithm}
\caption{Training Loop for \methodname}
\label{alg:growflow training}
\begin{algorithmic}[1]
    \State \textbf{Input:} Gaussians $\Gaussian$, posed images $I^t_{p}$, neural ODE $F_{\phi}$, number of timesteps $N$.
    \State \textbf{Parameters:} $n_{\text{static}} = 30000$, $n_{\text{boundary}} = 300$, $n_{\text{global}} = 30000$.
    
    \\ \State \textbf{Step 1: Static Reconstruction} \label{alg:phase1}
    \For{$epoch = 1$ to $n_{\text{static}}$} 
        \State Pick last timestep ground truth image $I_{\text{last}} = I_{p}^{T}$
        \State $I_{\text{pred}} \gets \text{Rasterize}(\Gaussian)$
        \State Compute $L \gets \text{loss}(I_{\text{pred}}, I_{\text{last}})$
        \State Update $\Gaussian$ 
    \EndFor 
    \State \textbf{Output:} $\Gaussian^{t_0} = (\centr^{t_0}, \quat^{t_0}, \scale^{t_0}, c, o)$
    
    \\ \State \textbf{Step 2: Boundary Reconstruction} \label{alg:phase2}
    \For{$k \in \{0, \ldots, N-1\}$} \Comment{Backwards in time}
        \For{$epoch = 1$ to $n_{\text{boundary}}$} 
        \State Pick ground truth image $I^{t_{k+1}}$
        \State $\Gaussian^{t_{k+1}} = \Gaussian^{t_k} + \int_{t_k}^{t_{k+1}} F_{\phi}(\centr(t), t)\, dt$ 
        \State $I_{\text{pred}} \gets \text{Rasterize}(\Gaussian^{t_{k+1}})$
        \State Compute $L \gets \text{loss}(I_{\text{pred}},I^{t_{k+1}})$
        \State Update $F_{\phi}$
        \EndFor
        \State Cache $\Gaussian^{t_{k+1}}$
    \EndFor
    \State \textbf{Output:} a set of cached Gaussians for each timestep $\{\Gaussian^{t_k}\}_k$
    
    \\ \State \textbf{Step 3: Global Optimization} \label{alg:phase3}
    \State Re-initialize new $F_{\phi}$
    \For{$epoch = 1$ to $n_{\text{global}}$} 
        \State Randomly sample timestep $t_k$
        \State Pick ground truth image $I^{t_{k+1}}$
        \State $\tilde{\Gaussian}^{t_{k+1}} = \Gaussian^{t_k} + \int_{t_k}^{t_{k+1}} F_{\phi}(\centr(t), t)\, dt$ 
        \State $I_{\text{pred}} \gets \text{Rasterize}(\tilde{\Gaussian}^{t_{k+1}})$
        \State Compute $L \gets \text{loss}(I_{\text{pred}},I^{t_{k+1}})$
        \State Update $F_{\phi}$
    \EndFor
    \State \textbf{Output:} Optimized $F_{\phi}$
\end{algorithmic}
\end{algorithm}
\section{Additional Visualizations}
\begin{figure*}
    \begin{center}
    \includegraphics[width=\linewidth]{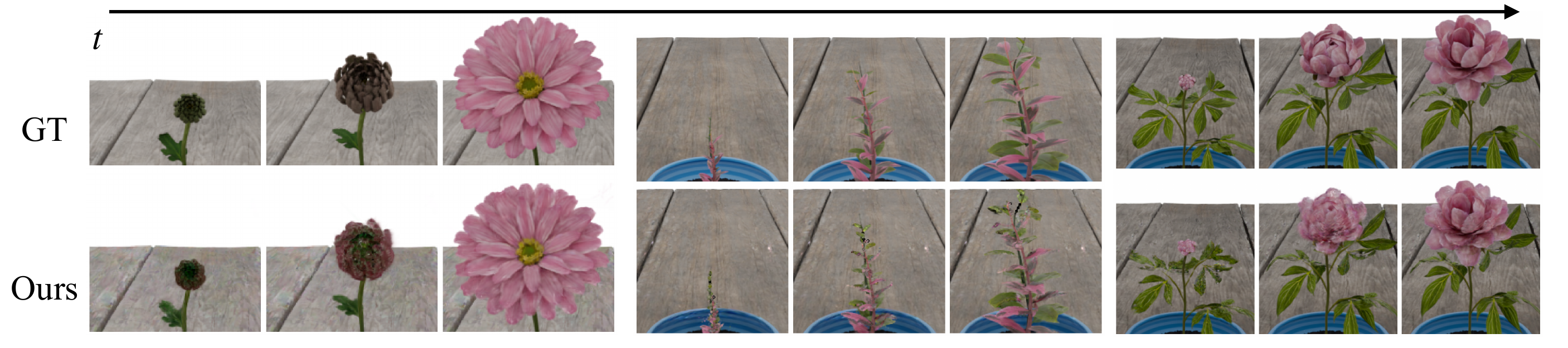}
    \end{center}
    \vspace{-6px}
    \caption{\textbf{Difficult scenes.} Our method also works on color-varying plants, multiple plant growth, and complex branching.}
    \label{fig:difficult_scene}
\end{figure*}
\paragraph{Adaptability to difficult scenes.} 
Our method can also reconstruct a variety of difficult plants such as color-varying plants, multiple plant growth, and complex branching (see \cref{fig:difficult_scene}). To model color-varying plants, we add an additional MLP, $\dot c = \mlp_{c}(\feat)$, integrated alongside other parameters.
\end{document}